\documentclass[letterpaper]{article} % DO NOT CHANGE THIS
\usepackage{aaai25}  % DO NOT CHANGE THIS
\usepackage{times}  % DO NOT CHANGE THIS
\usepackage{helvet}  % DO NOT CHANGE THIS
\usepackage{courier}  % DO NOT CHANGE THIS
\usepackage[hyphens]{url}  % DO NOT CHANGE THIS
\usepackage{graphicx} % DO NOT CHANGE THIS
\usepackage{natbib}  % DO NOT CHANGE THIS AND DO NOT ADD ANY OPTIONS TO IT
\usepackage{caption} % DO NOT CHANGE THIS AND DO NOT ADD ANY OPTIONS TO IT
\usepackage{algorithm}
\usepackage{algorithmicx}
\usepackage{algpseudocode}

\usepackage{amssymb,stmaryrd}

\usepackage{bm}
\usepackage{multirow}
\usepackage{booktabs}
\usepackage{array}
\usepackage{subfigure} 
\usepackage{amsthm}
\usepackage{amsmath}
\usepackage{threeparttable}
\usepackage{makecell}
\usepackage{longtable}

\usepackage{color}  % Basic support
\usepackage{xcolor} % Extended support

\usepackage{placeins}
\usepackage{afterpage}
\usepackage{float}

\usepackage{newfloat}
\usepackage{listings}
\DeclareCaptionStyle{ruled}{labelfont=normalfont,labelsep=colon,strut=off} % DO NOT CHANGE THIS
\floatstyle{ruled}
\newfloat{listing}{tb}{lst}{}
\floatname{listing}{Listing}

\title{Physics-Guided Foundation Model for Scientific Discovery: \\ An Application to Aquatic Science}
\author{
    Runlong Yu\textsuperscript{\rm 1}, 
    Chonghao Qiu\textsuperscript{\rm 1}, 
    Robert Ladwig\textsuperscript{\rm 2}, 
    Paul Hanson\textsuperscript{\rm 3}, 
    Yiqun Xie\textsuperscript{\rm 4}, 
    Xiaowei Jia\textsuperscript{\rm 1}
}
\affiliations{
    \textsuperscript{\rm 1}Department of Computer Science, University of Pittsburgh \\
    \textsuperscript{\rm 2}Department of Ecoscience, Aarhus University \\
    \textsuperscript{\rm 3}Center for Limnology, University of Wisconsin-Madison \\ 
    \textsuperscript{\rm 4}Department of Geographical Sciences, University of Maryland 
    \\

     \{ruy59,chq29,xiaowei\}@pitt.edu,
        rladwig@ecos.au.dk, pchanson@wisc.edu,  
        xie@umd.edu

}

\usepackage{bibentry}

\begin{document}

\maketitle

\begin{abstract}

Physics-guided machine learning (PGML) has become a prevalent approach in studying scientific systems due to its ability to integrate scientific theories for enhancing machine learning (ML) models. However, most PGML approaches are tailored to isolated and relatively simple tasks, which limits their applicability to complex systems involving multiple interacting processes and 
numerous influencing features. 
In this paper, we propose a \textit{\textbf{P}hysics-\textbf{G}uided \textbf{F}oundation \textbf{M}odel (\textbf{PGFM})} that combines pre-trained ML models and physics-based models and leverages their complementary strengths to improve the modeling of multiple coupled processes. 
To effectively conduct pre-training,  we construct a simulated environmental system that encompasses a wide range of influencing features and various simulated variables generated by physics-based models. The model is pre-trained in this system to adaptively select important feature interactions guided by multi-task objectives. 
We then fine-tune the model for each specific task using true observations, while maintaining consistency with established physical theories, such as the principles of mass and energy conservation. 
We demonstrate the effectiveness of this methodology in modeling water temperature and dissolved oxygen dynamics in real-world lakes. The proposed PGFM is also broadly applicable to a range of scientific fields 
where physics-based models are being used.

\end{abstract}

\section{Introduction}

Physics-based models of dynamical systems are often used to study scientific systems. For instance, scientists in aquatic science build physics-based models to simulate different water quality variables such as water temperature and dissolved oxygen (DO) concentrations, which are vital for assessing ecosystem health and water security. Similar models are also applied in agriculture~\cite{jia2019bringing}, geology~\cite{reichstein2019deep}, climate science~\cite{faghmous2014big}, and bio-medicine~\cite{yazdani2020systems}. 
Despite their widespread use, physics-based models face limitations due to the simplified representations of complex physical processes and the challenges of selecting appropriate parameters~\cite{jia2019physics}. Additional complexity arises when coupling these models to perform multiple tasks due to the dependencies amongst physical processes, e.g., DO concentrations are highly dependent on water temperature profiles. 

With advances in data collection driven by improved sensor technologies, there is a growing interest in using machine learning (ML) to extract complex data patterns for scientific problems ~\cite{willard2022integrating,he2023physics,wang2024simfair,xu2024spatial}.  
However, most ML approaches are only tested for isolated and simple tasks while still being limited in applicability to complex real-world systems with interacting and non-stationary processes.  
Recently, foundation models have shown great promise in tasks like vision and natural language processing by pre-training over large datasets~\cite{bommasani2021opportunities,zhou2023comprehensive,ye2024harnessing}. These models also offer tremendous opportunities for scientific modeling due to their ability to harness large and heterogeneous data and adapt to diverse downstream tasks. 

\begin{figure*} [!t]
\centering
	\includegraphics[width=0.69\linewidth]{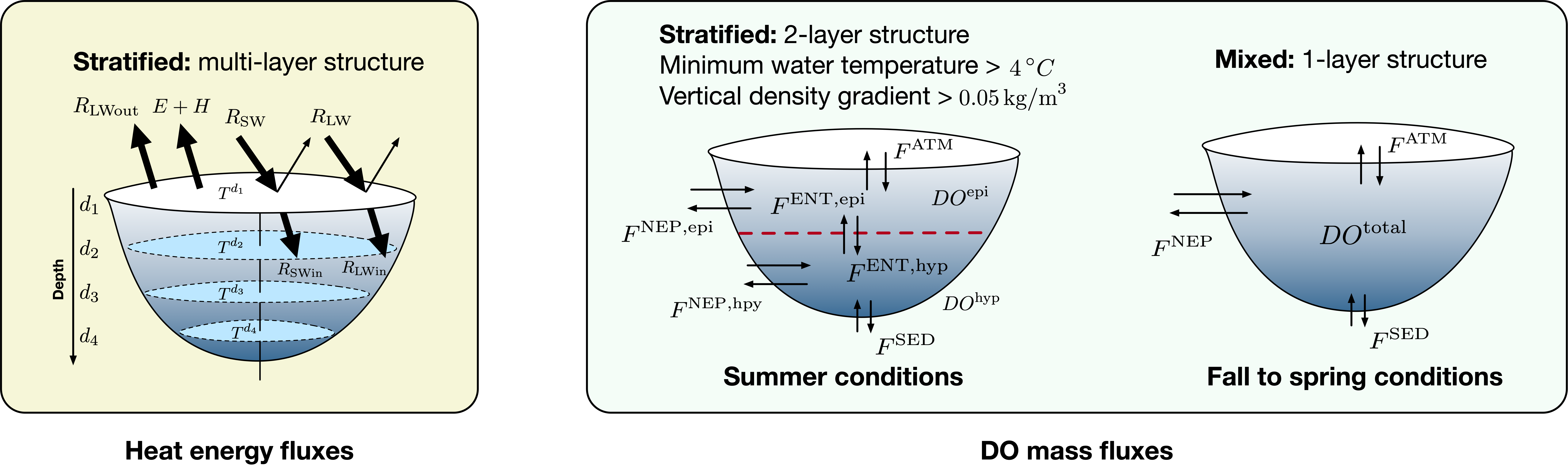}
	\caption{Heat energy and DO mass fluxes in the lake. } 
	\label{fig:1}
\end{figure*}

However, direct application of existing foundation models to scientific problems often leads to serious false discoveries due to several major challenges: \textit{1.~Data requirements}: Advanced ML models can often outperform traditional empirical models (e.g., regression), 
but these models require extensive training data, which is often scarce in real scientific applications. 
\textit{2.~Effectiveness of pre-training}:  Unsupervised pre-training has been shown to significantly boost the performance of many existing foundation models and mitigate the need for large data in downstream tasks. However, traditional pre-training tasks are not well aligned with scientific modeling tasks and thus can be less effective. The datasets used to pre-train existing models also have little overlap with target scientific data. 
\textit{3.~Physical consistency and generalizability}: 
With the absence of physical knowledge, existing foundation models can only learn statistical patterns from available data. Although the model could perform well in similar training data distribution, 
the patterns extracted by ML models may significantly violate some established physical relationships (e.g., mass and energy conservation). 
Consequently, these models can struggle to generalize to unseen scenarios. 
\textit{4.~Learning multiple tasks}: Most existing foundation models for scientific problems still focus on single prediction tasks~\cite{li2024lite,xie2024foundation} while largely ignoring the dependencies among multiple physical variables. 
This restricts their utility in complex systems characterized by multiple tasks and numerous influencing features.

To address these challenges, in this paper, we propose a \textit{\textbf{P}hysics-\textbf{G}uided \textbf{F}oundation \textbf{M}odel (\textbf{PGFM})} framework. 
Instead of relying on complex model architecture, PGFM integrates scientific knowledge from existing physics-based models to guide model pre-training and adaptation. Our proposed PGFM  is implemented and evaluated in the context of predicting water temperature and DO concentrations in lake systems. Both variables are key indicators of water quality and are intertwined with 
ecosystem phenology, influenced by features such as morphometric characteristics of lakes, weather conditions, trophic states, and watershed land use~\cite{read2019process,ladwig2022long,yu2024evolution}. 
Also, water temperature and DO concentrations are highly interdependent, with temperature shifts affecting oxygen solubility and biochemical reactions. Modeling the dynamics of these water quality variables can provide important insights for resource managers to make informed decisions to ensure safe drinking water, preserve aquatic habitat, and support sustainable water resource management.

Ideally, the foundation model should preserve three key properties. First, the model should be generalizable to large and diverse regions, e.g., predicting water quality in different lakes, even when they are sparsely observed. Second, the model should be able to perform multiple prediction tasks related to the target system. Third, the model needs to capture the physical dependencies between different tasks. 
To achieve these goals, 
the PGFM framework includes meticulously designed pre-training and fine-tuning stages.  In particular, the pre-training is conducted in a simulated environmental system that encompasses a wide range of data features and various simulated variables generated by physics-based models. By using an evolution-based algorithm, the foundation model progressively evolves to select features and their interactions that best reflect the dynamics of multiple simulated variables in the system. When fine-tuning the foundation model to each specific object (i.e., lake) and task (e.g., water temperature or DO concentration for a particular depth layer), we reuse the extracted features obtained from pre-training and also enhance the training objective with physical relationships (e.g., mass and energy conservation) specific to the task. To explicitly capture the interdependence between water temperature and DO concentration in this work, we augment the input for DO modeling with predicted lake temperatures, thereby enhancing the model's robustness by providing physically relevant information. 

Our evaluations on a wide range of lakes in the Midwestern USA demonstrate the capability of PGFM to effectively predict water temperature and DO concentration even with limited observed data. Our code is available at https://github.com/RunlongYu/PGFM.

% \textcolor{red}{Our evaluations on XXXXX demonstrate XXX}

\section{Problem Formulation}

The objective of this work is to predict daily water temperature profiles and DO concentrations at different depth layers in lake systems. 
As illustrated in Figure~\ref{fig:1}, thermal expansion properties of water facilitate stratification, creating a stable vertical density gradient. This results in distinct layers. Stratification inhibits vertical mixing, limiting the transfer of nutrients and oxygen between layers and reducing connectivity between the lower bottom and the atmosphere, thereby creating barriers to oxygen replenishment~\cite{read2011derivation}. To reflect summer changes in DO concentrations, our study focuses on stratified lakes with a vertical density difference exceeding $0.05 \, \text{kg/m}^3$ between surface and bottom layers, average water temperatures above $4\,^\circ \text{C}$, and a thermocline.

\begin{figure*}[!t]	\centerline{\includegraphics[width=0.82\linewidth]{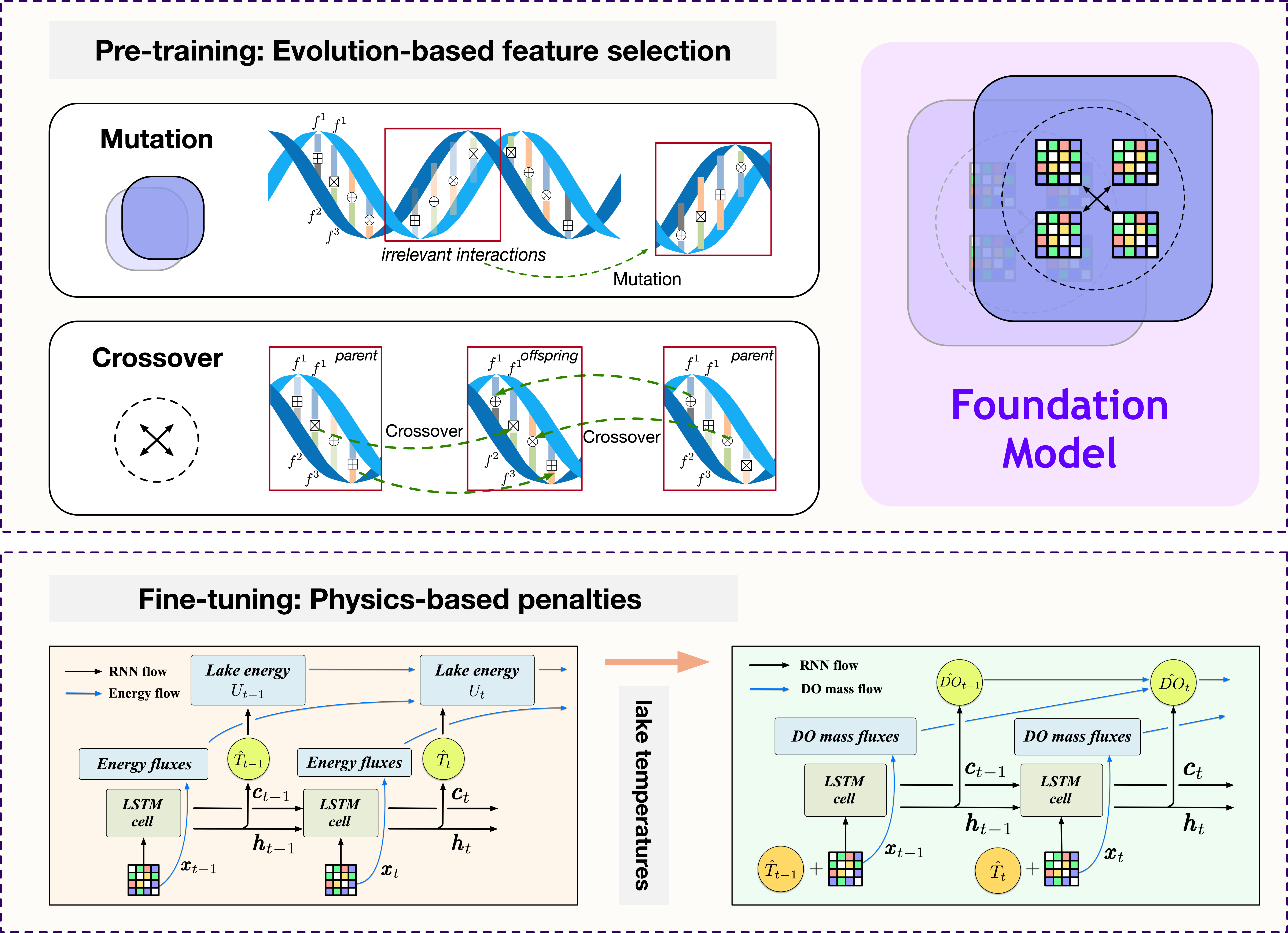}}
	\caption{ The overall framework of PGFM. }
	\label{fig：2}
\end{figure*}

More specifically, we aim to predict water temperature for multiple depth layers with an interval of $0.5 \, \text{m}$.
In contrast, we analyze the DO dynamics in two distinct depth layers of the water column: a well-mixed upper layer (epilimnion), and a cooler, nutrient-rich but light-limited deep layer (hypolimnion).
From fall to spring, when the water column is typically completely mixed, our model aims to predict the total DO concentration throughout the lake.

For each lake, we have access to its phenological features $\pmb{x}_{d,t}$ at each depth $d$ and time-step $t$. These features span a broad range of $m$ diverse fields, governing the dynamics of lake temperature and DO concentration, represented as $\pmb{x}_{d,t} = \{ x_{d,t}^1, \cdots, x_{d,t}^m \}$.
They include morphometric and geographic details such as lake area, depth, and shape; flux-related data like ecosystem and sedimentation fluxes; weather factors comprising wind speed and temperature; a range of trophic states from dystrophic to eutrophic; and diverse land use proportions extending from forests to wetlands. 
In addition to these input features, we observed water temperatures and DO concentrations on certain days and in certain depth layers. We use $T^d_t$ to represent the temperature at depth $d$ and time step $t$. During summer, we distinguish DO concentrations in the epilimnion, denoted as $DO_{t}^{\rm epi}$, from those in the hypolimnion, denoted as $DO_{t}^{\rm hyp}$. The thermocline, denoted by $tc$, determines which layer each measurement pertains to: $DO_{t}^{\rm epi}$ if the depth $d \leqslant tc$, and $DO_{t}^{\rm hyp}$ if $d > tc$. From fall to spring,  the recorded observations reflect the total DO concentration, denoted as $DO_{t}^{\rm total}$.

\section{Physics-Guided Foundation Model}

In this section, we introduce the Physics-Guided Foundation Model (PGFM) framework, as illustrated in Figure~\ref{fig：2}. The PGFM framework consists of two primary learning stages: (1) the pre-training stage and (2) the fine-tuning stage. 

\subsection{Pre-training Stage}

The success of existing foundation models is contingent upon effective pre-training overabundant representative data samples. Conventional pre-training tasks, e.g., masked token prediction, are not well aligned with scientific modeling tasks. This misalignment arises from the complex relationships between the input feature space (e.g., environmental conditions) and the target variable space (e.g., properties of lake systems). This challenge is further compounded by limited and less representative data. 
A key innovation of this work is pre-training in a physics-based simulated environmental system. This enables the development of a robust and generalizable model by leveraging comprehensive data that are well-aligned with established physical principles. The pre-training objective is to extract features and their interactions that are generalizable to various downstream tasks.  

In the following, we outline the simulated environmental system used in the pre-training stage, and then discuss the evolution-based feature selection that facilitates the learning and adaptation of the proposed foundation model.

\subsubsection{Simulated environmental system.} 
We first construct a simulated environmental system. It encompasses a wide range of data features and various simulated labels generated by physics-based models. For simulating lake temperatures,
we employ the general lake model (GLM)~\cite{hipsey2019general}, a widely used physics-based model that captures various heat energy fluxes affecting water temperature in lakes. These include the heating of the water surface from terrestrial long-wave radiation ($R_{\rm LW}$) and incoming short-wave radiation ($R_{\rm SW}$). The lake loses heat mainly through the outward fluxes of back radiation ($R_{\rm LWout}$), sensible heat fluxes ($H$), and latent evaporative heat fluxes ($E$), as illustrated in Fig.~\ref{fig:1}. For short-wave radiation ($R_{\rm SW}$) and long-wave radiation ($R_{\rm LW}$), a portion of the energy is reflected by the lake surface. 
For DO concentration dynamics, an advanced physics-based model is adopted,  as detailed in~\cite{ladwig2022long}.  The model simulates several key DO mass fluxes, including fluxes from the atmospheric exchange ($F^{\rm ATM}$), net ecosystem production ($F^{\rm NEP}$), and oxygen consumption by sediment ($F^{\rm SED}$). Additionally, during the summer, it accounts for DO entrainment fluxes from or into the other layer driven by turbulent flow ($F^{\rm ENT}$), as depicted in Fig.~\ref{fig:1}. Turbulent forces drive entrainment fluxes that either shallow or deepen the thermocline, affecting the transport of DO into either the epilimnion or the hypolimnion. 

\subsubsection{Evolution-based feature selection.} 
We train the foundation model to perform evolution-based feature selection in the simulated environmental system using a heuristic algorithm. 
This training process is conceived as an evolutionary search, akin to how organisms strive to evolve better traits for higher survival rates. In this analogy, we liken feature interactions to genomes and foundation models to organisms, where traits inherited via genes drive evolutionary success.

To facilitate this process, we use an embedding layer to convert input phenological features into a series of multi-field feature embeddings $\pmb{f}_{d,t}  = [\pmb{f}^1_{d,t}, \cdots, \pmb{f}^m_{d,t} ]$, where $\pmb{f}^i_{d,t} = \text{embed}(x^i_{d,t} )$. Using these embeddings, the aim of evolution-based feature selection can be formally described as identifying the most informative feature interactions to improve the prediction of target objectives, as
$\mathcal{H}: \mathcal{M}(\pmb{f}, \pmb{g}(\pmb{f}))\rightarrow \{\hat{T}, \hat{DO} \}$,  where $\pmb{g}$ denotes the set of operations to interact on feature pairs, and $\pmb{g}(\pmb{f})$ denotes the set of interactions. 
The algorithm $\mathcal{H}$ is designed to minimize the mean squared error loss $\mathcal{L}_{\rm FM}(\mathcal{M})$ for the outputs of the foundation model $\mathcal{M}$. The smaller the loss, the better the fitness of $\mathcal{M}$, reflecting a closer alignment between the predicted labels from the foundation model and the simulated labels from physics-based models. Here the simulated labels could include multiple variables involved in the aquatic systems, thus the loss $\mathcal{L}_{\rm FM}(\mathcal{M})$ can be a multi-task objective. 

To explicitly capture interactions amongst influencing features in the system, we introduce 
operations as the basic units of feature interaction. In particular, operations convert two individual features into interactions. Reflecting the diversity of genetic base pairs, we extend the operation set with four types of operations: $\pmb{g}=\{ \oplus, \otimes, \boxplus, \boxtimes \}$, which have been widely utilized in prior research~\cite{khawar2020autofeature,song2020towards,liu2020autogroup,yu2023cognitive}. As depicted in Fig~\ref{fig：2}, these operations encompass element-wise sum ($\oplus$), element-wise product ($\otimes$), and more complex forms like concatenation with a feed-forward layer ($\boxtimes$) and element-wise product with a feed-forward layer ($\boxplus$).

Motivated by the goal of enhancing model fitness through the preservation of beneficial genetic information, we aim to discern and prioritize important features and their interactions via a parameterized method. The idea is to introduce a set of relevance parameters to strengthen relevant feature interactions while diminishing or mutating those that contribute less.  
In this context, we define relevance parameters  for features $\pmb{f}_{d,t} $ and interactions $\tilde{g}(\pmb{f}_{d,t})$ as $\pmb{\alpha} = \{ \alpha_i | 1\leqslant i\leqslant m\} $ and $\pmb{\beta} =\{ \beta_{i,j} | 1\leqslant i<j\leqslant m \} $, respectively. Here, $\tilde{g}(\pmb{f}_{d,t})$ denotes the interaction of applying any operations from $\pmb{g}$ to a pair of features.
The predictive response of our model at time step $t$ is formulated as: $\mathcal{M}\big( \pmb{\alpha} \cdot \pmb{f}_{d,t},  \pmb{\beta} \cdot \tilde{g}(\pmb{f}_{d,t}) \big)$. Note that $\mathcal{M}$ is agnostic of specific ML-based models. In this work, we opt for long short-term memory (LSTM) networks~\shortcite{hochreiter1997long}, chosen for their proven effectiveness in capturing temporal dependencies in hydrology, as demonstrated in several studies~\cite{jia2021simlr,hanson2020predicting,chen2023physics}. We also test other advanced models (e.g., Transformer) in the experiments. 
We use a regularized dual averaging (RDA) optimizer to learn the relevance parameters $\pmb{\alpha}$ and $\pmb{\beta}$~\cite{xiao2009dual}, with the aim to distinguish between relevant and irrelevant feature interactions. When the absolute value of the cumulative gradient average value in a certain position in $\pmb{\alpha}$  or  $\pmb{\beta}$  is less than a threshold, the weight of that position in relevance parameters will be set to $0$, resulting in the sparsity of the relevance~\cite{xiao2009dual,liu2020autofis}. 

Mutation and crossover serve as key mechanisms of our evolution process. The mutation mechanism primarily aims at mutating the operations associated with irrelevant interactions into alternative operations, thus generating a new model (the offspring). 
For example, for an interaction $g_k(\pmb{f}^i_{d,t}, \pmb{f}^j_{d,t})$, mutation is triggered with a probability $\sigma$ after every $\tau$ steps if the relevance parameter $\beta_{i,j}$ drops below a threshold $\lambda$. In other words, 
to regenerate a new interaction, the operation $g_k$ of the interaction $g_k(\pmb{f}^i_{d,t}, \pmb{f}^j_{d,t})$ mutates into another operation $g_l $, which is randomly selected from the operation set as $g_l = \{ g\, | \, g \in \pmb{g}, g \neq g_k \}$. The new interaction $g_l(\pmb{f}^i_{d,t}, \pmb{f}^j_{d,t})$ replaces the irrelevant interaction $g_k(\pmb{f}^i_{d,t}, \pmb{f}^j_{d,t})$, and its corresponding relevance $\beta_{i,j}$ is reset. Consequently, the parent model $\mathcal{M}$ evolves into its offspring $\mathcal{M}'$.
The mutation mechanism is shown in Fig.~\ref{fig：2}. When the relevance of interactions is low (indicated by a lighter color), these are targeted for mutation, meaning that the operations of the interactions change into the other operations.

For a population-based search with a population size of $n$ ($n>1$), a crossover mechanism is used across multiple parent models to generate the offspring model. Consider $n$ random models as a population $\mathcal{P}$. For a model $\mathcal{M}_{\nu} \in \mathcal{P}$, we denote the relevance of features and interactions as $\pmb{\alpha}^{\mathcal{M}_{\nu}}$ and $\pmb{\beta}^{\mathcal{M}_{\nu}}$, respectively. Different models in the population may have various operations for the same feature pair $(f_i, f_j)$, represented as $g_{i,j}^{\mathcal{P}} = \{ g_{i,j}^{\mathcal{M}_1}, \cdots,  g_{i,j}^{\mathcal{M}_\nu}, \cdots,  g_{i,j}^{\mathcal{M}_n} \}$. We select the operation with the highest relevance for the offspring model, given as $g_{i,j}^{\mathcal{M}'} = \arg \max_{g_{i,j}^{\mathcal{M}_\nu} \in \, g_{i,j}^{\mathcal{P}}} \beta_{i,j}^{\mathcal{M}_{\nu}}$. We illustrate the crossover mechanism of two parents in Fig.~\ref{fig：2}. If the relevance of interactions of a parent is small (shown as lighter color), the operations should be selected from the other parents whose relevance of the interactions is large. Meanwhile, interactions of the offspring inherit their relevance from respective parents. 

\subsubsection{Instantiation of (n+1)-PGFM pre-training.} 
We present an instantiation to illustrate the steps for pre-training foundation models. In line with the canonical nomenclature used in evolution strategies, we refer to this as the (n+1)-PGFM.

Initially, (n+1)-PGFM creates a population of $n$ random models. For every $\tau$ iterations, the crossover mechanism generates an offspring from parent models, and mutation is applied to ensure diversity within the population. New parents are selected from both the parents and offspring, with offspring only advancing to the next generation's parent pool if its fitness meets or exceeds that of the least fit current parent, given as $\mathcal{M} = \arg \max _{\mathcal{M}_\nu \in \mathcal{P}} \mathcal{L}_{\rm FM}(\mathcal{M}_\nu)$. 
Additionally, the 1/5 successful rule is employed to adapt the search regions for the population, that is,  if previous iterations fail to improve the model significantly, it suggests that the model may be approaching a local optimum. In such cases, reducing the mutation probability can help exploit the promising region near the optimum more effectively~\shortcite{beyer2002evolution}.
Finally, the algorithm culminates by delivering the best foundation model in $\mathcal{P}$, given as $\mathcal{M} = \arg \min _{\mathcal{M}_\nu \in \mathcal{P}} \mathcal{L}_{\rm FM}(\mathcal{M}_\nu)$.

\subsubsection{Discussion and remark.}

We progressively evolve foundation models within the simulated environmental system, selecting important feature interactions that align with multi-task objectives. This approach offers two significant advantages. Firstly, it effectively mitigates the issue of limited observed labels in real-world environments. Secondly, by enabling the model to learn from extensive labels rooted in universal physical laws and diverse environments, the feature interactions identified by the foundation model demonstrate broad generality. This strategy addresses the constraints of traditional physics-guided machine learning, which typically focuses on isolated and simple scenarios.

\subsection{Fine-tuning Stage}

The fine-tuning process leverages the features and their interactions selected through the pre-training stage, refining them to capture the dynamics of specific target variables in a real system. 
It utilizes the observations of target variables as references while also regularizing the model with physical laws that govern the underlying processes. 

Specifically, we utilize standard ML training loss $\mathcal{L}_{\rm ML}$ that measures the difference between observed labels and predicted labels. Besides, we introduce the physical loss $\mathcal{L}_{\rm PHY}$, which measures the degree of violation of established physical laws such as energy or mass conservation. 
The fine-tuning loss function is formulated as: $\mathcal{L}_{\rm FT} = \mathcal{L}_{\rm ML} + \lambda_{\rm PHY} \mathcal{L}_{\rm PHY}$, where the hyper-parameter $\lambda_{\rm PHY}$ adjusts the balance between the standard ML loss and the physics-based penalties. 
It is noteworthy to mention that the computation of $\mathcal{L}_{\rm ML}$ relies on the sparsely available observations, and thus can only be defined on certain dates and depth layers when observations are available. In contrast, the physical loss does not require observed variables but only needs to check whether the predictions are consistent with known physical relationships. Hence, the physical loss can be applied to all the data points and thus contribute to learning better continuous dynamics. 
In the following, we detail the implementation of physical loss functions designed for modeling lake water temperatures and DO concentrations.

\subsubsection{Energy conservation loss. }
Fig.~\ref{fig:1} illustrates the major incoming and outgoing heat fluxes for a lake system.  
The impact on the balance between these fluxes 
results in changes to the lake's total thermal energy ($U_t$). 
Specifically, the relationship between the lake’s thermal energy $U_t$ and these energy fluxes should satisfy $\Delta U_t = R_{\rm SW} (1 - \alpha_{\rm SW}) + R_{\rm LWin} (1 - \alpha_{\rm LW}) - R_{\rm LWout} - E - H$, where $\Delta U_t = U_{t+1} - U_{t}$, 
$\alpha_{\rm SW}$ represents the short-wave albedo (the proportion of short-wave radiation reflected by the lake surface), and $\alpha_{\rm LW}$ represents the long-wave albedo. We denote the net gain of heat on the right side of the equation as $F_{\rm E}$.

In this context, we define the physical loss based on energy conservation, as 
$\mathcal{L}_{\rm PHY} = \sum_{t} {\rm ReLU}\, (|\Delta U_t - F_{\rm E}| - \tau_{\rm EC})$, where 
$\Delta U_t$ is computed directly as $U_{t+1} - U_{t}$, and $U_{t}$ is estimated as the volume-averaged water temperature predicted over different depth layers. 
The hyper-parameter $\tau_{\rm EC}$ represents a tolerance threshold for the violation of energy conservation. This is introduced to account for potential impacts from minor factors not included in estimating the heat fluxes or from observational errors in meteorological data. 
All the heat fluxes can be estimated from the input drivers. 

\begin{figure}
\centerline{\includegraphics[width=0.8\linewidth]{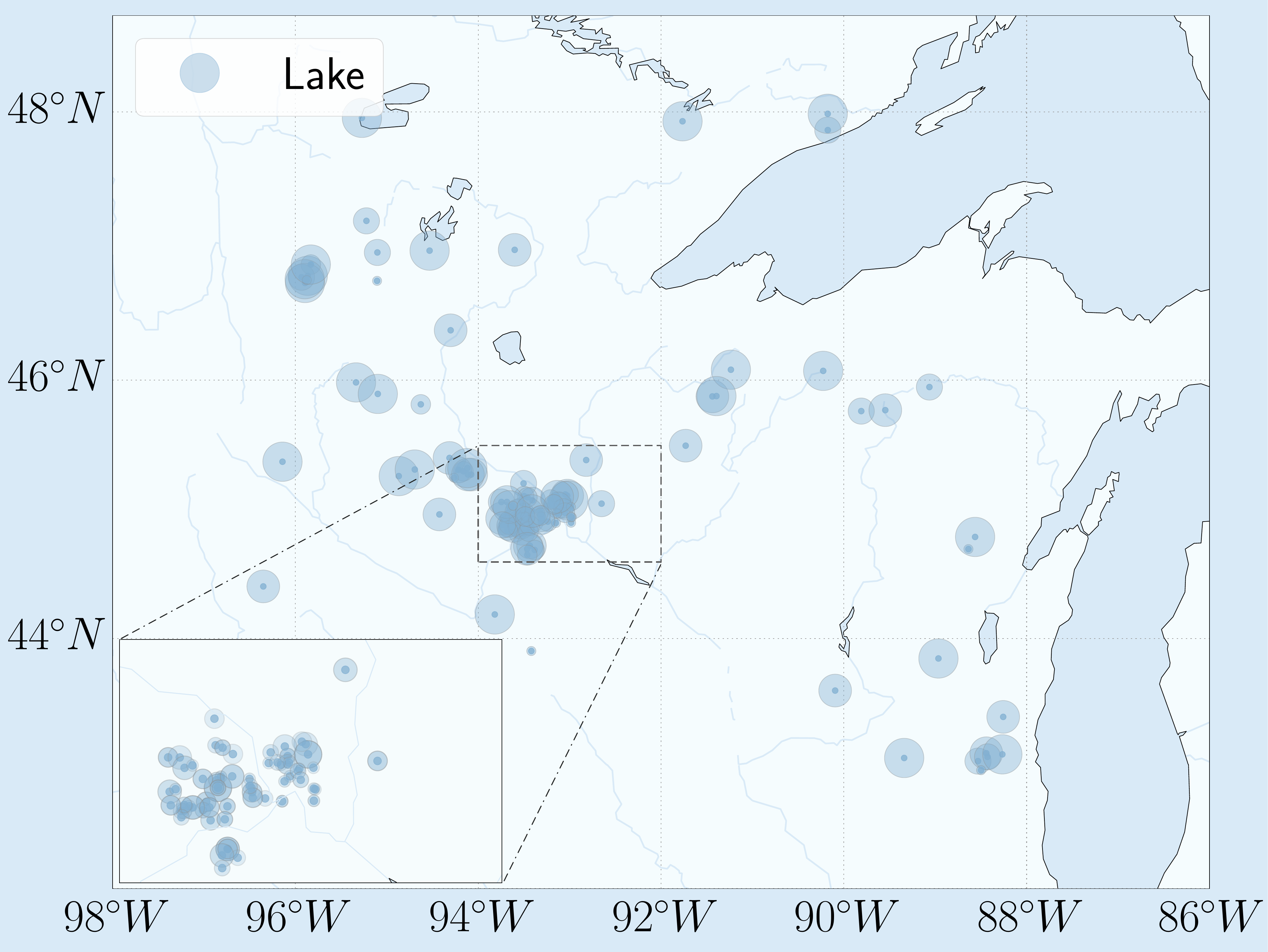}}
	\caption{Map of 117 tested lakes. } 
	\label{fig:3}
	\end{figure}

\newcolumntype{L}[1]{>{\raggedright\arraybackslash}p{#1}}
\newcolumntype{C}[1]{>{\centering\arraybackslash}p{#1}}
\newcolumntype{R}[1]{>{\raggedleft\arraybackslash}p{#1}}

		\begin{table*}[!t]
        \centering
        \begin{tabular}{L{3.2cm}C{2.2cm}C{2.2cm}C{2.45cm}C{2.5cm}C{2.4cm}}
			\toprule
			\multicolumn{1}{l}{\multirow{2}[3]{*}{Algo. Name}} & \multicolumn{2}{c}{Water temperature ($^\circ C$) } & \multicolumn{3}{c}{DO concentration  ($g / m^{3}$)} \\
			\cmidrule(lr){2-3} \cmidrule(lr){4-6} 
			& Summer & Fall to spring & Summer (epi.) & Summer (hyp.) & Fall to spring \\
			\midrule
			Phy-based & 2.795 \textcolor{gray}{(0.000)} & 1.624 \textcolor{gray}{(0.000)} & 
			2.277 \textcolor{gray}{(0.000)} & 2.367 \textcolor{gray}{(0.000)} & 2.481 \textcolor{gray}{(0.000)}  \\
			LSTM & 3.841 \textcolor{gray}{(0.290)} & 3.929 \textcolor{gray}{(0.498)} & 
			2.825 \textcolor{gray}{(0.160)} & 2.775 \textcolor{gray}{(0.166)} & 2.908 \textcolor{gray}{(0.313)}  \\
			EA-LSTM & 4.590 \textcolor{gray}{(0.378)} & 2.993 \textcolor{gray}{(0.575)} & 
			3.936 \textcolor{gray}{(0.160)} & 3.759 \textcolor{gray}{(0.165)} & 4.654 \textcolor{gray}{(0.279)}  \\
			Transformer & 3.589 \textcolor{gray}{(0.606)} & 4.806 \textcolor{gray}{(1.283)} & 
			2.678 \textcolor{gray}{(0.374)} & 2.625 \textcolor{gray}{(0.320)} & 3.148 \textcolor{gray}{(0.638)} \\
            iTransformer & 2.828 \textcolor{gray}{(0.231)} & 2.619 \textcolor{gray}{(0.313)} & 
			3.137 \textcolor{gray}{(1.021)} & 3.127 \textcolor{gray}{(0.424)} & 2.433 \textcolor{gray}{(1.021)} \\
			\midrule
            LSTM (w/ pre-train) & 2.249 \textcolor{gray}{(0.229)} & 1.779 \textcolor{gray}{(0.543)} & 
			2.306 \textcolor{gray}{(0.309)} & 2.389 \textcolor{gray}{(0.304)} & 2.414 \textcolor{gray}{(0.784)}  \\
			FM+LSTM & 2.003 \textcolor{gray}{(0.011)} & 1.578 \textcolor{gray}{(0.030)} & 2.117 \textcolor{gray}{(0.012)} & 2.292 \textcolor{gray}{(0.008)} & 2.350 \textcolor{gray}{(0.017)}  \\
			FM+Transformer & 2.177 \textcolor{gray}{(0.197)} & 1.603 \textcolor{gray}{(0.334)} & 
			2.145 \textcolor{gray}{(0.200)} & 2.346 \textcolor{gray}{(0.201)} & 2.264 \textcolor{gray}{(0.352)}  \\
			\midrule
            PGFM (w/o pre-train) & 3.578 \textcolor{gray}{(0.938)} & 3.044 \textcolor{gray}{(0.939)} & 2.772 \textcolor{gray}{(0.242)} & 2.725 \textcolor{gray}{(0.226)} & 2.833 \textcolor{gray}{(0.291)}  \\
			PGFM (w/o $\hat{T}$) & --- & --- & 
			2.104 \textcolor{gray}{(0.126)} & 2.170 \textcolor{gray}{(0.106)} & 2.267 \textcolor{gray}{(0.311)}  \\
			PGFM & \textbf{1.953} \textcolor{gray}{(0.126)} & \textbf{1.365} \textcolor{gray}{(0.200)} & 
			\textbf{2.077} \textcolor{gray}{(0.111)} & \textbf{2.162} \textcolor{gray}{(0.114)} & \textbf{2.258} \textcolor{gray}{(0.288)}  \\
			\bottomrule
            \end{tabular}%
              \caption{Comparative performance in predicting water temperature and DO concentration in terms of RMSE.}
              \label{tab:1}
		\end{table*}%

\subsubsection{DO mass conservation loss. } 
Referring back to Fig.~\ref{fig:1}, we categorize the fluxes caused by atmospheric exchange ($F^{\rm ATM}$), net ecosystem production ($F^{\rm NEP}$), and mineralization through sediment oxygen demand ($F^{\rm SED}$), among other factors, as exogenous fluxes ($F^{\rm EXO}$).
In the well-mixed conditions from fall to spring, assuming that diurnal variations in total lake volume are negligible, we can model the DO dynamics as $\tilde{DO}_{t}^{\rm total} = \hat{DO}_{t-1}^{\rm total} + F^{\rm EXO}_{t-1} \times \Delta t$. During the stratified conditions typical of summer, the dynamics become more complex. It becomes necessary to account for daily volume changes in both the epilimnion and hypolimnion, as well as the entrainment fluxes between layers caused by turbulent flow ($F^{\rm ENT}$). The modeling is adjusted accordingly:
{\small$ \tilde{DO}_{t}^{\rm epi} = \left( \hat{DO}_{t-1}^{\rm epi} + F^{\rm EXO,epi}_{t-1} \times \Delta t\right) \times \frac{V_{t-1}^{\rm epi}}{V_t^{\rm epi}} + F^{\rm ENT,epi}_{t-1}
$} and {\small$ \tilde{DO}_{t}^{\rm hyp} = \left( \hat{DO}_{t-1}^{\rm hyp} + F^{\rm EXO, hyp}_{t-1} \times \Delta t\right) \times \frac{V_{t-1}^{\rm hyp}}{V_t^{\rm hyp}} + F^{\rm ENT, hyp}_{t-1} $}, where $V^{\rm epi}_t$ and $V^{\rm hyp}_t$ represent the volumes of the epilimnion and hypolimnion, respectively. 
In the context of predicting DO concentrations, we define the physical loss as $\mathcal{L}_{\rm PHY} = \sum_{t} {\rm ReLU}\, ( | \hat{DO}_{t} - \tilde{DO}_t | -\tau_{\rm MC} ) $, with $\tau_{\rm MC}$ set as a tolerance threshold for the mass conservation loss. 

\begin{figure} [!t]
\centering
	\includegraphics[width=1\linewidth]{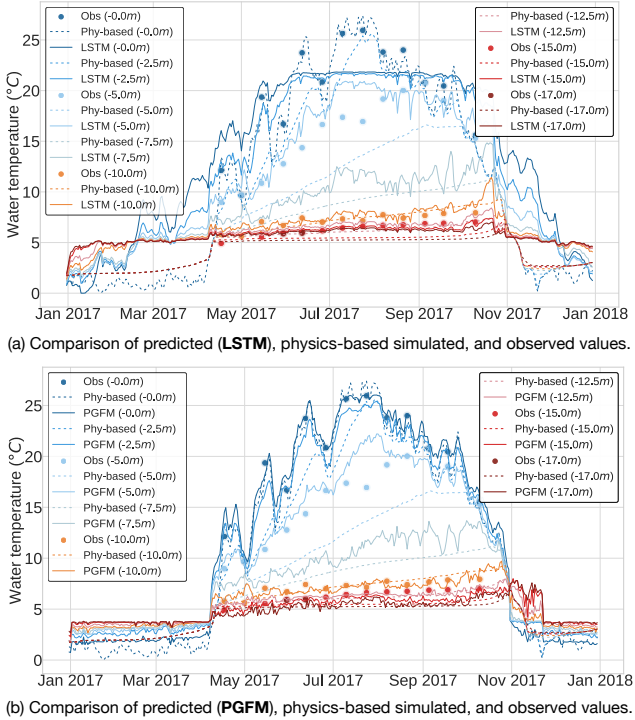}
	\caption{Time-series analysis of water temperature. } 
	\label{fig:4}
\end{figure}

Recognizing the interdependence between these tasks, particularly how temperature fluctuations influence oxygen solubility and biochemical reactions, we substitute the temperature from the simulated environmental system in the input variable $\pmb{x}$ with the predicted lake temperature $\hat{T}_t$ for predicting DO concentrations. This change enhances the accuracy and relevance of the model's predictions. 
Additionally, exogenous flux ($F^{\rm EXO}$) and lake volume ($V_t$) are included in the input variables $\pmb{x}$. The entrainment fluxes ($F^{\rm ENT}$) are calculated based on the predicted DO concentration ($\hat{DO}_{t}$) and fluctuations of the thermocline ($tc$).

\subsubsection{Discussion and remark.}

The integration of the physical loss $\mathcal{L}_{\rm PHY}$, based on energy conservation and mass conservation, 
offers several benefits. By aligning the machine learning model with established physical principles, this approach effectively narrows the search space, enhancing model performance, particularly in scenarios with sparse data and out-of-sample conditions. Moreover, the computation of  $\mathcal{L}_{\rm PHY}$ does not require observed values and thus can be implemented on large unlabeled data points.

\section{Experimental Evaluation} 
\label{sec:exp}

\subsubsection{Data preparation. } 

We evaluate the proposed PGFM framework for predicting water temperature and DO concentration using a comprehensive dataset covering 41 years (1979–2019). This dataset includes ecological observations from 117 lakes in the Midwestern USA, as shown in Figure~\ref{fig:3}. The color intensity of each point indicates lake depth, while the size represents surface area. The dataset consists of approximately 1.75 million daily records, each with 47 phenological features such as morphometric attributes, weather conditions, trophic states, and land use.  
Data source descriptions are available in~\cite{meyer2024national,yu2024adaptive,willard2021predicting}. Of these, 57,156 days contain 476,215 observed water temperature measurements (across depths), and 23,192 days include observed DO concentrations in epilimnion and hypolimnion during summer or total DO concentrations under mixed conditions from fall to spring.

\begin{figure} [!t]
\centering
	\includegraphics[width=1\linewidth]{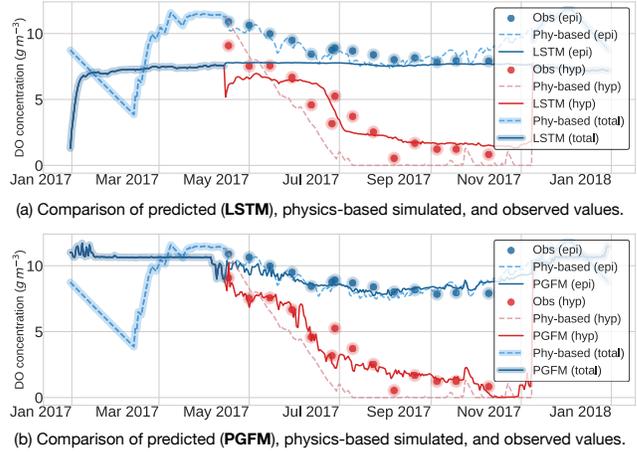}
	\caption{Time-series analysis of DO concentrations. } 
	\label{fig:5}
\end{figure} 

\begin{figure*} [!t]
	\centering
	\includegraphics[width=0.89\linewidth]{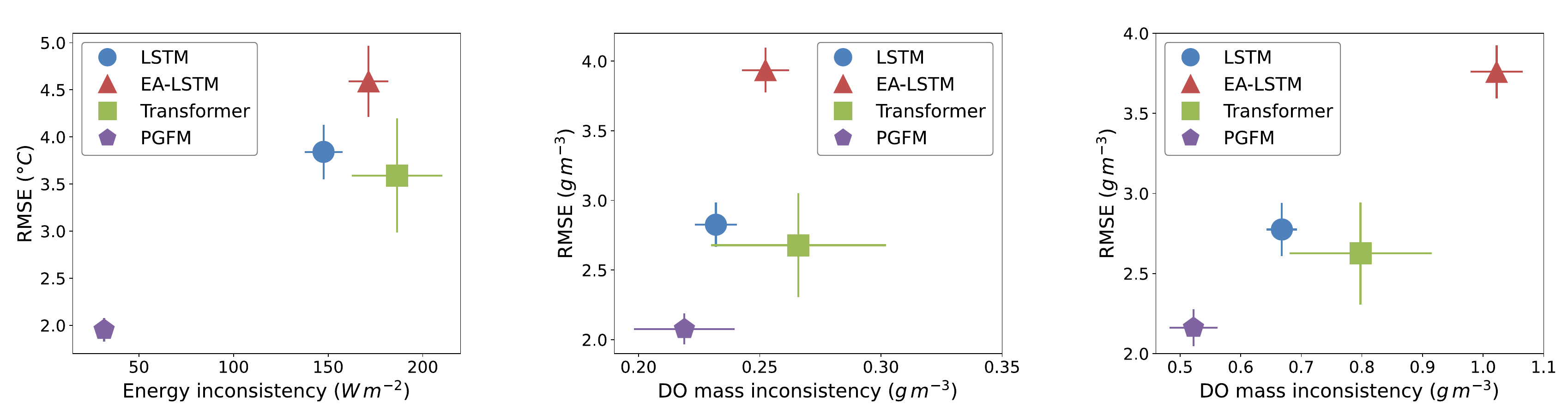}
	\caption{Physical consistency analysis: water temperature (left), epilimnion (center) and hypolimnion (right) DO concentration.} 
	\label{fig:6}
\end{figure*}

\subsubsection{Baselines.} 

To demonstrate our PGFM's effectiveness, we compare it against several baselines, including task-specific physics-based models~\cite{hipsey2019general,ladwig2022long}, LSTM~\shortcite{hochreiter1997long}, EA-LSTM~\cite{kratzert2019towards}, Transformer~\cite{vaswani2017attention}, and iTransformer~\cite{liuitransformer}. Among these, we establish LSTM (w/ pre-train) as a baseline, which is simply pre-trained on simulated data. To evaluate the performance of our proposed foundation model, we fine-tune the pre-trained foundation model using LSTM and Transformer, denoted as FM+LSTM and FM+Transformer, respectively. These models are fine-tuned without incorporating physics-based penalties. In contrast, our PGFM approach integrates LSTM with physics-based penalties during the fine-tuning phase, ensuring better adherence to physical principles. We also evaluate a variant of PGFM that skips the pre-training stage but retains physics-based penalties, referred to as PGFM (w/o pre-train). Additionally, to assess the impact of incorporating predicted temperatures on DO prediction, we evaluate a version of PGFM without the inclusion of predicted temperatures on the DO prediction task, designated as PGFM (w/o $\hat{T}$).

\subsubsection{Performance comparison (RQ1).}
Table~1 presents a comparative analysis of PGFM against baseline methods for predicting water temperature and DO concentration, with evaluations tailored to the distinct mixing conditions of water bodies between summer and fall to spring. Water temperature evaluations average prediction errors across all layers, while DO concentration is assessed separately for the epilimnion and hypolimnion layers in summer, and as total DO concentration under mixed conditions from fall to spring. Performance is measured using root mean square error (RMSE), with results including both the mean and standard deviation (indicated in grey) from five runs. 

From the results, we observe the following key insights:
First, physics-based models generally outperform ML models, primarily due to the limited availability of observed data, which hinders the generalization ability of ML models to unseen conditions.
Second, pre-training with our proposed foundation model significantly improves the performance of both LSTM and Transformer, surpassing physics-based models and LSTM (w/ pre-train). This improvement is due to the foundation model’s ability to leverage extensive labels rooted in universal physical laws and diverse environments, allowing it to identify broadly applicable feature interactions.
Third, unlike simply pre-training an LSTM or omitting the pre-training stage—both of which risk overlooking complex, nonlinear relationships—the evolutionary algorithm identifies critical feature interactions that enhance predictive accuracy and physical consistency. Fourth, direct comparisons of PGFM with baseline methods show substantial performance gains across all evaluated periods, demonstrating its effectiveness in both tasks.
Lastly, incorporating predicted water temperature into the DO concentration predictions during the fine-tuning stage further enhances performance, underscoring the value of integrating the related task to improve overall model efficacy.

\begin{figure} [!t]
	\centering
	\includegraphics[width=0.9\linewidth]{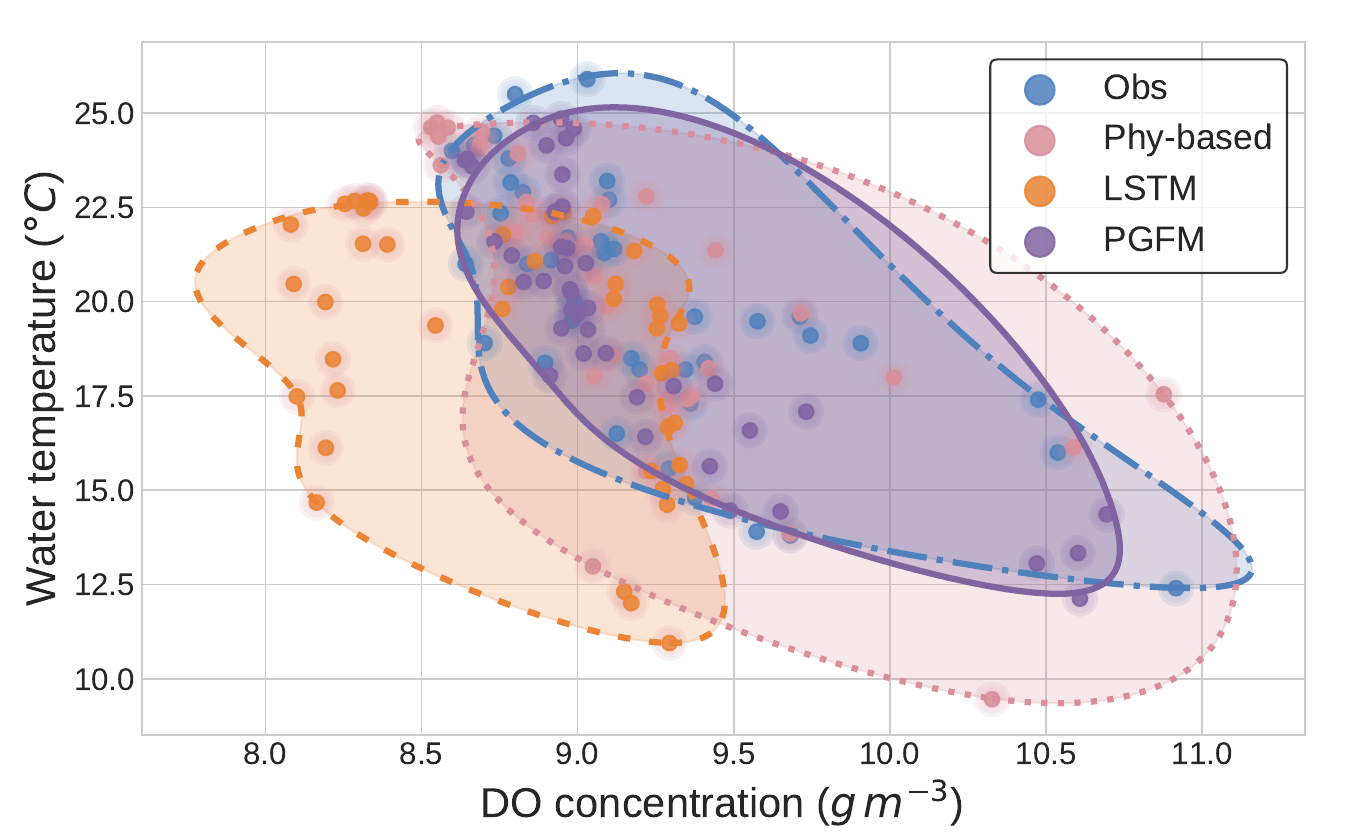}
	\caption{Surface water temperature and epilimnion DO concentrations: predictions vs. observations for the same period. } 
	\label{fig:7}
\end{figure}

\subsubsection{Time-series analysis (RQ2).} Figure~\ref{fig:4} provides a time-series comparison of water temperature predictions from LSTM and PGFM against physics-based simulations and observed values at several depth layers. Figure~\ref{fig:5} does the same for DO concentration. These comparisons specifically highlight results from the summer season of the testing period, given the limited availability of observed data from fall to spring and the heightened concern for DO in lakes during summer (when the hypolimnion often experiences oxygen depletion, potentially leading to aquatic organism fatalities).

The figures illustrate a sharp drop in lake temperature at a certain depth, indicative of the thermocline, and distinct DO patterns in the epilimnion and hypolimnion layers during the stratified period. The analysis reveals that PGFM not only aligns more closely with observed values compared to LSTM but also captures subtle fluctuations more effectively, demonstrating its sensitivity and ability to accurately track trends seen in physics-based simulations. In contrast, LSTM struggles to capture these critical dynamics, making its results less reliable. Additionally, it is evident that physics-based methods generally struggle with accurately predicting lake bottom temperature and DO concentration.

\begin{figure} [!t]
	\centering
	\includegraphics[width=0.75\linewidth]{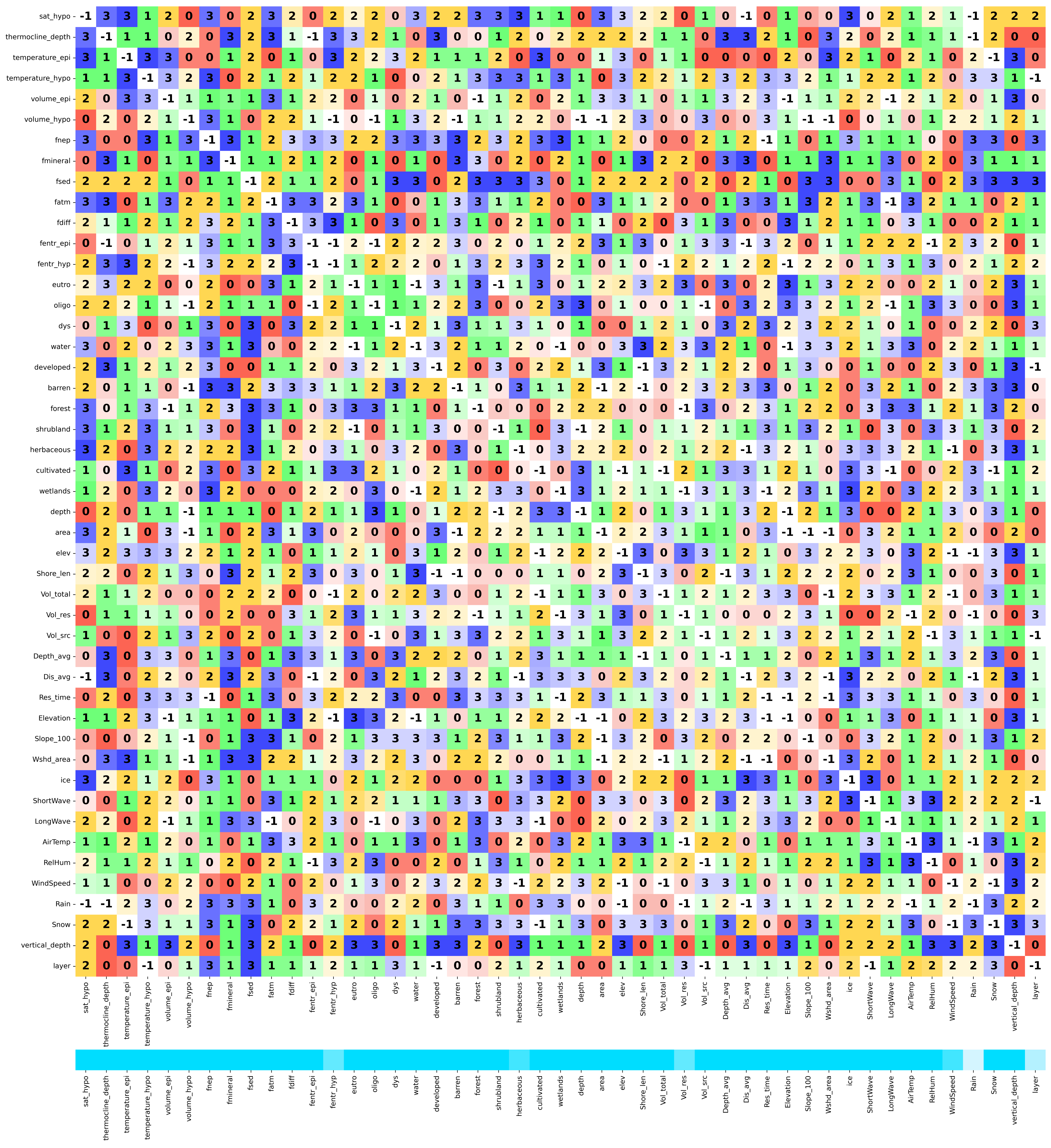}
	\caption{Gene map of PGFM. } 
	\label{fig:8}
\end{figure}

\subsubsection{Physical consistency analysis (RQ3).} 
In scientific applications, machine learning models are expected to align with observed data and maintain physical consistency. To demonstrate how PGFM enhances physical consistency, Figure~\ref{fig:6} displays the RMSE and physical inconsistency metrics (i.e., energy and mass inconsistency) for each method's predictions of water temperature (left), epilimnion DO concentrations (center), and hypolimnion DO concentrations (right) during summer conditions. Physics-based models are excluded from this analysis as they inherently exhibit zero physical inconsistency. PGFM consistently positions closest to the bottom left corner, underscoring its superior ability to reduce both prediction RMSE and physical inconsistency.

Physical consistency is also evident in the relationships between different variables, such as the well-known principle that oxygen solubility decreases as water temperature increases.
Figure~\ref{fig:7} showcases predictions from various methods alongside observed values of surface water temperature and epilimnion DO concentrations for a lake during the same period. Analysis of the envelope curves reveals that PGFM closely matches the observed values and most accurately reflects this physical principle. In contrast, LSTM fails to capture the relationship between these two variables accurately.

\subsubsection{Selected feature interactions (RQ4).}
To demonstrate the evolutionary process of PGFM and how feature interactions evolve under multi-task guidance, we visualize the \textbf{gene maps} of PGFM in Figure~\ref{fig:8}. Using an encoding where $\oplus=0, \otimes=1,  \boxplus=2, \boxtimes=3$, we represent the model's fitness as a symmetric matrix. Distinct colors are allocated to each operation, creating a vibrant gene map where each gene symbolizes an interaction; like red ``$0$'', green ``$1$'', yellow ``$2$'', and blue ``$3$''. For example, a green ``$1$'' within the ``\verb|depth| $\times$ \verb|area|'' block signifies that the element-wise product~$\otimes$ is identified as the optimal operation for ``\verb|depth|'' to interact with ``\verb|area|''. The color intensity on the gene map correlates with the relevance of the interactions, where darker hues signify higher relevance and lighter hues indicate lesser importance. Each feature is also visually represented by uniformly colored bars. Interactions deemed irrelevant, with their relevance parameters reduced to $0$, are omitted, leaving their corresponding genes depicted in white ``$-1$''. One observation from the analysis is that water temperature and DO concentration are predominantly influenced by features such as water volume, weather conditions, and air temperature, with relatively minor effects from local land use factors.

\section{Conclusion} 
This paper proposed a Physics-Guided Foundation Model (PGFM) for scientific discovery. PGFM leverages a wide range of influencing features and various simulated variables generated by physics-based models for pre-training, enabling it to learn from extensive labels rooted in universal physical laws and diverse environments. We applied the PGFM framework specifically to aquatic science, where we developed physical loss functions based on principles of energy and mass conservation and incorporated them into the fine-tuning stage. 
In the future, we encourage the adaptation of this idea to other scientific fields to explore its potential.

\section*{Acknowledgments}

This work was supported by the National Science Foundation (NSF) under grants 2239175, 2316305, 2213549, 2126474, 2147195, 2425844, 2425845, and 2430978, the USGS awards  G21AC10564 and G22AC00266, and the NASA grant 80NSSC24K1061. Yiqun Xie gratefully acknowledges the support of Google’s AI for Social Good Impact Scholars program. This research was also supported in part by the University of Pittsburgh Center for Research Computing through the resources provided. 

We also acknowledge the data contributions from the U.S. Geological Survey, NASA, the HydroSHEDS project led by the World Wildlife Fund, and other collaborative institutions for providing essential datasets on water temperature, lake characteristics, trophic states, and land use patterns.

 \small{

}

%\small{\bibliography{mybibliography}}
%\small{\bibliography{anonymous-submission-latex-2025-v2}}

\begin{thebibliography}{36}
\providecommand{\natexlab}[1]{#1}

\bibitem[{Beyer and Schwefel(2002)}]{beyer2002evolution}
Beyer, H.-G.; and Schwefel, H.-P. 2002.
\newblock Evolution strategies--a comprehensive introduction.
\newblock \emph{Natural Computing}, 1: 3--52.

\bibitem[{Bommasani et~al.(2021)Bommasani, Hudson, Adeli, Altman, Arora, von
  Arx, Bernstein, Bohg, Bosselut, Brunskill
  et~al.}]{bommasani2021opportunities}
Bommasani, R.; Hudson, D.~A.; Adeli, E.; Altman, R.; Arora, S.; von Arx, S.;
  Bernstein, M.~S.; Bohg, J.; Bosselut, A.; Brunskill, E.; et~al. 2021.
\newblock On the opportunities and risks of foundation models.
\newblock \emph{arXiv preprint arXiv:2108.07258}.

\bibitem[{Chen et~al.(2023)Chen, Xie, Li, Liang, and Jia}]{chen2023physics}
Chen, S.; Xie, Y.; Li, X.; Liang, X.; and Jia, X. 2023.
\newblock Physics-guided meta-learning method in baseflow prediction over large
  regions.
\newblock In \emph{Proceedings of the 2023 SIAM International Conference on
  Data Mining (SDM)}, 217--225. SIAM.

\bibitem[{Faghmous and Kumar(2014)}]{faghmous2014big}
Faghmous, J.~H.; and Kumar, V. 2014.
\newblock A big data guide to understanding climate change: The case for
  theory-guided data science.
\newblock \emph{Big data}, 2(3): 155--163.

\bibitem[{Hanson et~al.(2020)Hanson, Stillman, Jia
  et~al.}]{hanson2020predicting}
Hanson, P.~C.; Stillman, A.~B.; Jia, X.; et~al. 2020.
\newblock Predicting lake surface water phosphorus dynamics using
  process-guided machine learning.
\newblock \emph{Ecological Modelling}, 430: 109136.

\bibitem[{He et~al.(2023)He, Xie, Liu, Chen, Jin, and Jia}]{he2023physics}
He, E.; Xie, Y.; Liu, L.; Chen, W.; Jin, Z.; and Jia, X. 2023.
\newblock Physics guided neural networks for time-aware fairness: an
  application in crop yield prediction.
\newblock In \emph{Proceedings of the AAAI Conference on Artificial
  Intelligence}, volume~37, 14223--14231.

\bibitem[{Hipsey, Bruce et~al.(2019)}]{hipsey2019general}
Hipsey, M.~R.; Bruce, L.~C.; et~al. 2019.
\newblock A General Lake Model (GLM 3.0) for linking with high-frequency sensor
  data from the Global Lake Ecological Observatory Network (GLEON).
\newblock \emph{Geoscientific Model Development}.

\bibitem[{Hochreiter and Schmidhuber(1997)}]{hochreiter1997long}
Hochreiter, S.; and Schmidhuber, J. 1997.
\newblock Long short-term memory.
\newblock \emph{Neural computation}, 9(8): 1735--1780.

\bibitem[{Jia et~al.(2019{\natexlab{a}})Jia, Khandelwal, Mulla, Pardey, and
  Kumar}]{jia2019bringing}
Jia, X.; Khandelwal, A.; Mulla, D.~J.; Pardey, P.~G.; and Kumar, V.
  2019{\natexlab{a}}.
\newblock Bringing automated, remote-sensed, machine learning methods to
  monitoring crop landscapes at scale.
\newblock \emph{Agricultural Economics}, 50: 41--50.

\bibitem[{Jia et~al.(2019{\natexlab{b}})Jia, Willard, Karpatne, Read, Zwart,
  Steinbach, and Kumar}]{jia2019physics}
Jia, X.; Willard, J.; Karpatne, A.; Read, J.; Zwart, J.; Steinbach, M.; and
  Kumar, V. 2019{\natexlab{b}}.
\newblock Physics guided RNNs for modeling dynamical systems: A case study in
  simulating lake temperature profiles.
\newblock In \emph{Proceedings of the 2019 SIAM international conference on
  data mining}, 558--566. SIAM.

\bibitem[{Jia et~al.(2021)Jia, Xie, Li, Chen, Zwart, Sadler, Appling, Oliver,
  and Read}]{jia2021simlr}
Jia, X.; Xie, Y.; Li, S.; Chen, S.; Zwart, J.; Sadler, J.; Appling, A.; Oliver,
  S.; and Read, J. 2021.
\newblock Physics-guided machine learning from simulation data: An application
  in modeling lake and river systems.
\newblock In \emph{2021 IEEE International Conference on Data Mining (ICDM)},
  270--279. IEEE.

\bibitem[{Khawar et~al.(2020)Khawar, Hang, Tang, Liu, Li, and
  He}]{khawar2020autofeature}
Khawar, F.; Hang, X.; Tang, R.; Liu, B.; Li, Z.; and He, X. 2020.
\newblock Autofeature: Searching for feature interactions and their
  architectures for click-through rate prediction.
\newblock In \emph{ACM International Conference on Information and Knowledge
  Management (CIKM)}, 625--634.

\bibitem[{Kratzert et~al.(2019)}]{kratzert2019towards}
Kratzert, F.; et~al. 2019.
\newblock Towards learning universal, regional, and local hydrological
  behaviors via machine learning applied to large-sample datasets.
\newblock \emph{Hydrology and Earth System Sciences}, 23.

\bibitem[{Ladwig et~al.(2022)Ladwig, Appling, Delany, Dugan, Gao, Lottig,
  Stachelek, and Hanson}]{ladwig2022long}
Ladwig, R.; Appling, A.~P.; Delany, A.; Dugan, H.~A.; Gao, Q.; Lottig, N.;
  Stachelek, J.; and Hanson, P.~C. 2022.
\newblock Long-term change in metabolism phenology in north temperate lakes.
\newblock \emph{Limnology and Oceanography}, 67(7): 1502--1521.

\bibitem[{Li et~al.(2024)Li, Liu, Wang, Luo, Jia, and Yao}]{li2024lite}
Li, H.; Liu, J.; Wang, Z.; Luo, S.; Jia, X.; and Yao, H. 2024.
\newblock LITE: Modeling Environmental Ecosystems with Multimodal Large
  Language Models.
\newblock \emph{arXiv preprint arXiv:2404.01165}.

\bibitem[{Liu et~al.(2020{\natexlab{a}})Liu, Xue, Guo, Tang, Zafeiriou, He, and
  Li}]{liu2020autogroup}
Liu, B.; Xue, N.; Guo, H.; Tang, R.; Zafeiriou, S.; He, X.; and Li, Z.
  2020{\natexlab{a}}.
\newblock AutoGroup: Automatic feature grouping for modelling explicit
  high-order feature interactions in CTR prediction.
\newblock In \emph{Proceedings of the 43rd international ACM SIGIR conference
  on Research and Development in Information Retrieval (SIGIR)}, 199--208.

\bibitem[{Liu et~al.(2020{\natexlab{b}})Liu, Zhu, Li, Zhang
  et~al.}]{liu2020autofis}
Liu, B.; Zhu, C.; Li, G.; Zhang, W.; et~al. 2020{\natexlab{b}}.
\newblock Autofis: Automatic feature interaction selection in factorization
  models for click-through rate prediction.
\newblock In \emph{Proceedings of the 26th ACM SIGKDD International Conference
  on Knowledge Discovery \& Data Mining (KDD)}, 2636--2645.

\bibitem[{Liu et~al.(2024)Liu, Hu, Zhang, Wu, Wang, Ma, and
  Long}]{liuitransformer}
Liu, Y.; Hu, T.; Zhang, H.; Wu, H.; Wang, S.; Ma, L.; and Long, M. 2024.
\newblock iTransformer: Inverted Transformers Are Effective for Time Series
  Forecasting.
\newblock In \emph{The Twelfth International Conference on Learning
  Representations}.

\bibitem[{Meyer et~al.(2024)Meyer, Topp, King, Ladwig, Pilla, Dugan, Eggleston,
  Hampton, Leech, Oleksy et~al.}]{meyer2024national}
Meyer, M.~F.; Topp, S.~N.; King, T.~V.; Ladwig, R.; Pilla, R.~M.; Dugan, H.~A.;
  Eggleston, J.~R.; Hampton, S.~E.; Leech, D.~M.; Oleksy, I.~A.; et~al. 2024.
\newblock National-scale remotely sensed lake trophic state from 1984 through
  2020.
\newblock \emph{Scientific Data}, 11(1): 77.

\bibitem[{Read et~al.(2011)Read, Hamilton, Jones, Muraoka, Winslow, Kroiss, Wu,
  and Gaiser}]{read2011derivation}
Read, J.~S.; Hamilton, D.~P.; Jones, I.~D.; Muraoka, K.; Winslow, L.~A.;
  Kroiss, R.; Wu, C.~H.; and Gaiser, E. 2011.
\newblock Derivation of lake mixing and stratification indices from
  high-resolution lake buoy data.
\newblock \emph{Environmental Modelling \& Software}, 26(11): 1325--1336.

\bibitem[{Read et~al.(2019)Read, Jia, Willard, Appling, Zwart, Oliver,
  Karpatne, Hansen, Hanson, Watkins et~al.}]{read2019process}
Read, J.~S.; Jia, X.; Willard, J.; Appling, A.~P.; Zwart, J.~A.; Oliver, S.~K.;
  Karpatne, A.; Hansen, G.~J.; Hanson, P.~C.; Watkins, W.; et~al. 2019.
\newblock Process-guided deep learning predictions of lake water temperature.
\newblock \emph{Water Resources Research}, 55(11): 9173--9190.

\bibitem[{Reichstein et~al.(2019)Reichstein, Camps-Valls, Stevens, Jung,
  Denzler, Carvalhais, and Prabhat}]{reichstein2019deep}
Reichstein, M.; Camps-Valls, G.; Stevens, B.; Jung, M.; Denzler, J.;
  Carvalhais, N.; and Prabhat, F. 2019.
\newblock Deep learning and process understanding for data-driven Earth system
  science.
\newblock \emph{Nature}, 566(7743): 195--204.

\bibitem[{Song et~al.(2020)Song, Cheng, Zhou, Yang, Tian, and
  Hu}]{song2020towards}
Song, Q.; Cheng, D.; Zhou, H.; Yang, J.; Tian, Y.; and Hu, X. 2020.
\newblock Towards automated neural interaction discovery for click-through rate
  prediction.
\newblock In \emph{Proceedings of the 26th ACM SIGKDD International Conference
  on Knowledge Discovery \& Data Mining (KDD)}, 945--955.

\bibitem[{Vaswani(2017)}]{vaswani2017attention}
Vaswani, A. 2017.
\newblock Attention is all you need.
\newblock \emph{Advances in Neural Information Processing Systems}.

\bibitem[{Wang et~al.(2024)Wang, Xie, Li, Jia, Jiang, Jia, and
  Xu}]{wang2024simfair}
Wang, Z.; Xie, Y.; Li, Z.; Jia, X.; Jiang, Z.; Jia, A.; and Xu, S. 2024.
\newblock SimFair: Physics-Guided Fairness-Aware Learning with Simulation
  Models.
\newblock In \emph{Proceedings of the AAAI Conference on Artificial
  Intelligence}, volume~38, 22420--22428.

\bibitem[{Willard et~al.(2022)Willard, Jia, Xu, Steinbach, and
  Kumar}]{willard2022integrating}
Willard, J.; Jia, X.; Xu, S.; Steinbach, M.; and Kumar, V. 2022.
\newblock Integrating scientific knowledge with machine learning for
  engineering and environmental systems.
\newblock \emph{ACM Computing Surveys}, 55(4): 1--37.

\bibitem[{Willard et~al.(2021)Willard, Read, Appling, Oliver, Jia, and
  Kumar}]{willard2021predicting}
Willard, J.~D.; Read, J.~S.; Appling, A.~P.; Oliver, S.~K.; Jia, X.; and Kumar,
  V. 2021.
\newblock Predicting Water Temperature Dynamics of Unmonitored Lakes With
  Meta-Transfer Learning.
\newblock \emph{Water Resources Research}.

\bibitem[{Xiao(2009)}]{xiao2009dual}
Xiao, L. 2009.
\newblock Dual averaging method for regularized stochastic learning and online
  optimization.
\newblock \emph{Advances in Neural Information Processing Systems}, 22.

\bibitem[{Xie et~al.(2024)Xie, Wang, Chen, Li, Jia, Li, Wang, Chai, Li, and
  Skakun}]{xie2024foundation}
Xie, Y.; Wang, Z.; Chen, W.; Li, Z.; Jia, X.; Li, Y.; Wang, R.; Chai, K.; Li,
  R.; and Skakun, S. 2024.
\newblock When are Foundation Models Effective? Understanding the Suitability
  for Pixel-Level Classification Using Multispectral Imagery.
\newblock \emph{arXiv preprint arXiv:2404.11797}.

\bibitem[{Xu et~al.(2024)Xu, Xiao, He, Wang, Jiang, Chen, Xie, Jia, Yan, and
  Zhou}]{xu2024spatial}
Xu, Z.; Xiao, T.; He, W.; Wang, Y.; Jiang, Z.; Chen, S.; Xie, Y.; Jia, X.; Yan,
  D.; and Zhou, Y. 2024.
\newblock Spatial-Logic-Aware Weakly Supervised Learning for Flood Mapping on
  Earth Imagery.
\newblock In \emph{Proceedings of the AAAI Conference on Artificial
  Intelligence}, volume~38, 22457--22465.

\bibitem[{Yazdani et~al.(2020)Yazdani, Lu, Raissi, and
  Karniadakis}]{yazdani2020systems}
Yazdani, A.; Lu, L.; Raissi, M.; and Karniadakis, G.~E. 2020.
\newblock Systems biology informed deep learning for inferring parameters and
  hidden dynamics.
\newblock \emph{PLoS computational biology}, 16(11): e1007575.

\bibitem[{Ye et~al.(2024)Ye, Zheng, Shen, Wang, Zhang, Zhu, Yu, Zhang, and
  Xiong}]{ye2024harnessing}
Ye, Y.; Zheng, Z.; Shen, Y.; Wang, T.; Zhang, H.; Zhu, P.; Yu, R.; Zhang, K.;
  and Xiong, H. 2024.
\newblock Harnessing multimodal large language models for multimodal sequential
  recommendation.
\newblock \emph{arXiv preprint arXiv:2408.09698}.

\bibitem[{Yu et~al.(2024{\natexlab{a}})Yu, Ladwig, Xu, Zhu, Hanson, Xie, and
  Jia}]{yu2024evolution}
Yu, R.; Ladwig, R.; Xu, X.; Zhu, P.; Hanson, P.~C.; Xie, Y.; and Jia, X.
  2024{\natexlab{a}}.
\newblock Evolution-Based Feature Selection for Predicting Dissolved Oxygen
  Concentrations in Lakes.
\newblock In \emph{International Conference on Parallel Problem Solving from
  Nature}, 398--415. Springer.

\bibitem[{Yu et~al.(2024{\natexlab{b}})Yu, Qiu, Ladwig, Hanson, Xie, Li, and
  Jia}]{yu2024adaptive}
Yu, R.; Qiu, C.; Ladwig, R.; Hanson, P.~C.; Xie, Y.; Li, Y.; and Jia, X.
  2024{\natexlab{b}}.
\newblock Adaptive Process-Guided Learning: An Application in Predicting Lake
  DO Concentrations.
\newblock In \emph{2024 IEEE International Conference on Data Mining (ICDM)},
  580--589. IEEE.

\bibitem[{Yu et~al.(2023)Yu, Xu, Ye, Liu, and Chen}]{yu2023cognitive}
Yu, R.; Xu, X.; Ye, Y.; Liu, Q.; and Chen, E. 2023.
\newblock Cognitive Evolutionary Search to Select Feature Interactions for
  Click-Through Rate Prediction.
\newblock In \emph{Proceedings of the 29th ACM SIGKDD Conference on Knowledge
  Discovery and Data Mining}, 3151--3161.

\bibitem[{Zhou et~al.(2023)Zhou, Li, Li, Yu, Liu, Wang, Zhang, Ji, Yan, He
  et~al.}]{zhou2023comprehensive}
Zhou, C.; Li, Q.; Li, C.; Yu, J.; Liu, Y.; Wang, G.; Zhang, K.; Ji, C.; Yan,
  Q.; He, L.; et~al. 2023.
\newblock A comprehensive survey on pretrained foundation models: A history
  from bert to chatgpt.
\newblock \emph{arXiv preprint arXiv:2302.09419}.

\end{thebibliography}

\end{document}